\documentclass[letterpaper]{article} 
\usepackage{aaai24}  
\usepackage{times}  
\usepackage{helvet}  
\usepackage{courier}  
\usepackage[hyphens]{url}  
\usepackage{graphicx} 
\usepackage{natbib}  
\usepackage{caption} 
\usepackage{algorithm}
\usepackage{algorithmic}
\usepackage{amsfonts}
\usepackage{amsmath}
\usepackage{amssymb}
\usepackage[table]{xcolor}

\usepackage{algorithmic}
\usepackage{graphicx}
\usepackage{textcomp}
\usepackage{xcolor}
\def\BibTeX{{\rm B\kern-.05em{\sc i\kern-.025em b}\kern-.08em
    T\kern-.1667em\lower.7ex\hbox{E}\kern-.125emX}}

\usepackage{tikz}
\usepackage{graphicx,multirow}

\DeclareRobustCommand{\check}{%
  \tikz\fill[scale=0.4, color=green]
  (0,.35) -- (.25,0) -- (1,.7) -- (.25,.15) -- cycle;%
}

\newcommand{\cross}{%
\tikz[scale=0.23] {
    \draw[line width=0.7,line cap=round, color=red] (0,0) to [bend left=6] (1,1);
    \draw[line width=0.7,line cap=round, color=red] (0.2,0.95) to [bend right=3] (0.8,0.05);
}}

\usepackage{lineno}

\usepackage{newfloat}
\usepackage{listings}
\DeclareCaptionStyle{ruled}{labelfont=normalfont,labelsep=colon,strut=off} 
\floatstyle{ruled}
\newfloat{listing}{tb}{lst}{}
\floatname{listing}{Listing}
\title{Towards Financially Inclusive Credit Products Through Financial Time Series Clustering}
\author {
    Tristan Bester,
    Benjamin Rosman
}
\affiliations {
    School of Computer Science and Applied Mathematics \\
    University of the Witwatersrand \\
    Johannesburg, South Africa\\
    tristanbester@gmail.com, benjamin.rosman1@wits.ac.za
}

\begin{document}

\maketitle

\begin{abstract}
Financial inclusion ensures that individuals have access to financial products and services that meet their needs. As a key contributing factor to economic growth and investment opportunity, financial inclusion increases consumer spending and consequently business development. It has been shown that institutions are more profitable when they provide marginalised social groups access to financial services. Customer segmentation based on consumer transaction data is a well-known strategy used to promote financial inclusion. While the required data is available to modern institutions, the challenge remains that segment annotations are usually difficult and/or expensive to obtain. This prevents the usage of time series classification models for customer segmentation based on domain expert knowledge. As a result, clustering is an attractive alternative to partition customers into homogeneous groups based on the spending behaviour encoded within their transaction data. In this paper, we present a solution to one of the key challenges preventing modern financial institutions from providing financially inclusive credit, savings and insurance products: the inability to understand consumer financial behaviour, and hence risk, without the introduction of restrictive conventional credit scoring techniques. We present a novel time series clustering algorithm that allows institutions to understand the financial behaviour of their customers. This enables unique product offerings to be provided based on the needs of the customer, without reliance on restrictive credit practices.
\end{abstract}

\section{Introduction}
Financial inclusion means that individuals have access to financial products and services that meet their needs. Notably, financial inclusion is a strong contributing factor to economic growth as it stimulates entrepreneurship while expanding investment opportunities. It boosts consumer spending and business development, leading to job creation and improved productivity. Financial institutions are strongly incentivised to support the positive social impact brought about by a financially inclusive market \cite{moin2012use}. This is a consequence of the fact that institutions are more profitable when they make use of modern technological strategies that provide marginalised social groups access to financial services \cite{alshehadeh2022financial}.

A well-known strategy used to enable financial inclusion is customer segmentation. A key result in service marketing research indicates that companies should not market the same set of services to all customers, but rather target products at specific segments of the customer base \cite{ansari2016customer}. Customer segmentation can be achieved through the analysis of bank transaction data, which has become available as a result of the large-scale automated data acquisition systems present in modern financial institutions \cite{taifi2023intertwined}. Unfortunately, the challenge remains that data annotations are difficult and/or expensive to obtain from domain experts. This lack of annotations prevents supervised learning models from being used to perform customer segmentation based on domain expert knowledge \cite{dempster2020rocket, ismail2019deep}. As a result, unsupervised clustering is an attractive alternative to partition customers into homogeneous groups based on the spending behaviours encoded within their transaction data.

In this paper, we present a solution to one of the key challenges preventing modern financial institutions from providing financially inclusive credit, savings and insurance products. This is the inability to understand consumer financial behaviour, and hence risk, without the introduction of restrictive conventional credit scoring techniques. To this end, we extend the evaluation of time series clustering algorithms by measuring their performance on financial data. Guided by these results, we present a novel approach to financial time series clustering that is shown to outperform state-of-the-art techniques. Our method enables institutions to understand their customer base as a set of homogeneous groups, each with similar financial behaviour. By understanding the group to which a customer belongs, a product offering can be tailored to suit the needs of the customer in accordance with the risk appetite of the institution.

\section{Background}
\subsection{Clustering}
Cluster analysis or clustering is the task of partitioning a set of objects such that the elements within each partition are more similar to one another than those in other partitions. The notion of similarity in the space is defined both in terms of the distance metric as well as the algorithm responsible for the calculation of the clustering. As a result, cluster analysis is not comprised of a single algorithm, but rather a collection of techniques. The definition of what constitutes a cluster differs significantly between algorithms, and consequently, the approaches followed to efficiently calculate them. 

Cluster analysis is an unsupervised learning technique. Consequently, the ideal cluster associated with each sample in the dataset is unknown. This aspect of clustering makes it difficult to quantify the quality of a given clustering. This follows from the fact that the information required to compute such a quantity directly is not available. This has led to the introduction of proxy objectives such as the \textit{Silhouette Coefficient (SC)} and \textit{Davies-Bouldin index (DBI)} which measure the quality of clusterings based on assumed properties of high-quality clusterings. The SC quantifies the degree of similarity between elements within each cluster in comparison to that of other clusters. The SC ranges from -1 to 1, where a high value indicates that the objected is well suited to its cluster and poorly suited to neighbouring clusters. The DBI is a measure of similarity between the elements of a cluster and those of the closest neighbouring cluster. The similarity is defined as the ratio of within-cluster distances to between-cluster distances. The minimum DBI value is zero with lower values indicating better clusterings. 

\subsection{Specificity of the Time Dimension}
In this paper we focus on \textit{time series} data, where a time series is comprised of a collection of data points that are indexed through time. Each data point is a vector of feature values at a specific instant in time. It has been shown that conventional clustering algorithms perform poorly on this type of data, as they make use of distance metrics that are incompatible with the time series representation \cite{lafabregue2022end}. In response, several dedicated time series clustering algorithms have been developed. These algorithms either make use of distance metrics specifically designed for time series \cite{muller2007dynamic} or introduce alternative data representations \cite{madiraju2018deep}. The latter approach, known as \textit{representation learning}, is the primary focus of this study.

\subsection{Neural Network Architectures}
The architecture of a neural network refers to the number, types and sizes of the layers from which it is composed. Throughout the literature, several variations have been proposed for deep neural network layers. These layers can be organised into three families: \textit{fully connected, convolutional} and \textit{recurrent layers}.

\subsubsection{Fully Connected Neural Network Layers (FCNN)}
Each neuron in a fully connected layer is connected to all neurons in the following layer \cite{block1962analysis}. Fully connected layers apply a linear transformation to the input data before passing the result through a non-linear activation function. 

\subsubsection{Convolutional Neural Network Layers (CNN)}
Convolutional neural network layers are well known as a result of their widespread adoption in the field of computer vision \cite{he2016deep}. A key property of the convolutional kernels used within each layer is shift-invariance. This property is a consequence of the shared-weight architecture on which the design is based. While the most commonly used convolutional layers, designed for image processing, make use of two-dimensional convolutions, one-dimensional variants exist \cite{wang2017time}. These variants have been adapted to capture temporal rather than spatial patterns in the data.

\subsubsection{Recurrent Neural Network Layers (RNN)}
The recurrent neural network layer has been specifically designed for applications involving the time dimension \cite{hopfield1982neural}. Recurrent layers introduce cyclical connections, allowing neuron outputs to propagate through the inference process, subsequently affecting future inputs to the layer. These layers can effectively capture temporal dynamic behaviour. An RNN layer variant known as the \textit{Long Short-Term Memory (LSTM)} layer has been proposed in \cite{hochreiter1997long} as a solution to the vanishing gradient problem which hinders the performance of RNNs.

\subsection{Deep Representation Learning}
The objective of representation learning is to produce an alternate representation of the raw data. This representation is intended to increase the likelihood of obtaining desirable clustering results when used in conjunction with clustering algorithms that have not been designed for time series data. Representation learning is a key component of most deep learning-based clustering frameworks. Generally, the primary role of deep learning within these frameworks is to produce an alternate representation of the data. This is achieved through the use of a deep neural network known as the encoder. The encoder is a non-linear mapping $E_{\phi}: \mathcal{X} \xrightarrow{} \mathcal{Z}$, parameterised by $\phi$, that maps elements from the original data space $\mathcal{X}$ to the space of encoded representations $\mathcal{Z}$. The encoded representation of $x \in \mathcal{X}$, denoted $z = E_{\phi}(x)$, is referred to as the latent representation. The objective of deep representation learning is to learn a mapping $E_{\phi}$ that can produce a latent representation of the data that facilitates the desired clustering when used in combination with clustering algorithms that have not been designed for time series data.

We seek to optimise the parameters of the encoder such that the learned representation favours the clustering of data points into homogeneous groups. In this study, such a clustering corresponds to one in which individuals with similar transaction data, measured in terms of size and frequency of transaction amounts, are clustered together. However, as the desired clusters are unknown, we must optimise the encoder through the introduction of an auxiliary objective. Consequently, a second neural network known as the decoder is introduced along with a self-supervised objective function. Similarly to the encoder, the decoder $D_{\theta}: \mathcal{Z} \rightarrow \mathcal{X}$ is a non-linear mapping, parameterised by $\theta$, that maps elements from the latent space $\mathcal{Z}$ back to the original data space $\mathcal{X}$. The encoder and decoder functions form what is known as an autoencoder. Forward propagation in an autoencoder consists of two stages. Firstly, the input feature vector $x$ is embedded into the latent space through the use of the encoder network. Secondly, a reconstruction of $x$, denoted $x'$, is produced by passing the latent representation through the decoder network. The introduction of the decoder network facilitates the usage of a self-supervised objective. 

\subsubsection{Autoencoder Architectures} 
The architecture of an autoencoder consists of two main components, the encoder and decoder. As described, the encoder and decoder are both neural networks which function as non-linear mappings between vector spaces. In general, the decoder is constructed as a mirror of the encoder with the exception of the embedding layer \cite{lafabregue2022end}. That is, the architecture of the decoder is obtained by arranging the layers of the encoder in reverse order. Depending on the application, any of the previously described neural network layers can form part of an autoencoder's architecture.

\subsubsection{Autoencoder Training}
As described, the objective of representation learning is to produce an alternate representation of the raw data that facilitates the desired clustering. This requires the parameters of the encoder to be optimised towards this objective. As the desired cluster labels are unavailable, preventing the usage of supervised objectives, many auxiliary objectives have been used throughout the literature. While several approaches have been developed to train the parameters of the encoder, we focus on those that rely on the tasks of data reconstruction and generation. 

In the first approach to autoencoder training, the parameters of the encoder and decoder networks are optimised such that the network can encode and decode data while incurring minimal loss of information. More precisely, the network is trained to minimise the error between the input and reconstructed representations with respect to some loss function. The most commonly used objective function for this task is the classical reconstruction loss \cite{meng2017relational}. The classical reconstruction loss is defined as the mean squared error between the original and reconstructed representations:
\begin{equation} \label{eq:1}
    \mathcal{L}_{\mathcal{R}} = \frac{1}{N} \sum\limits_{i=1}^{N} \vert\vert x_i - D_{\theta}(E_{\phi}(x_i)) \vert\vert_2^2
\end{equation}
This loss was extended by \citet{ghasedi2017deep} in which a layerwise objective is introduced. That is, the loss is defined as the sum of reconstruction losses associated with each depth in the autoencoder network:
\begin{equation} \label{eq:2}
    \mathcal{L}_{\mathcal{LR}} = \frac{1}{N} \sum\limits_{i=1}^{N} \sum\limits_{l=1}^{L} \frac{1}{\vert z_i^l \vert} (z_i^l - \hat{z}_i^l)^2
\end{equation}
where $L$ is the number of encoder and decoder layers, $z_i^l$ is the output of the $l^{th}$ encoder layer, $\hat{z}_i^l$ is the output of the $l^{th}$ decoder layer and $\vert z_i^l \vert$ is the number of elements in the output of the $l^{th}$ encoder layer.

In contrast to reconstructive approaches, generative methods optimise the parameters of the encoder for the generation of realistic data. The \textit{variational autoencoder (VAE)} introduced by \citet{kingma2013auto} is one of the most commonly used generative methods. While this method is based on the autoencoder architecture, a significantly different training regime is introduced for parameter optimisation. The input is passed through the encoder in which it is mapped onto a Gaussian distribution $q_{\phi}(z|x)$. Samples are then drawn from this distribution and passed through the decoder to obtain the distribution $p_{\theta}(x|z)$. The parameters of the generative model $D_{\theta}$ (the decoder) are optimised to reduce the reconstruction error between the input and output representations. The parameters of the encoder network $E_{\phi}$ are optimised to minimise the distance between the two distributions $q_{\phi}(z|x)$ and $p_{\theta}(z|x)$. The proposed loss function to perform this task is known as the \textit{evidence lower bound (ELBO)}:
\begin{equation} \label{eq:3}
    \mathcal{L}_{\mathcal{V}} = \sum \limits_{i=1}^{N}
    \underset{z \sim q_{\phi}(\cdot|x_i)}{\mathbb{E}}
    \ln
    \bigg(
    \frac{p_\theta(x_i, z)}{q_\phi(z | x_i)}
    \bigg)
\end{equation}

\subsubsection{Encoder Optimisation for Clustering}
The aforementioned optimisation tasks have focused on the development of a meaningful latent space for specific tasks. While this approach provides good performance in some cases \cite{barkhordar2021clustering}, there is no guarantee that the learned representation will be well suited to the clustering task. In response to this challenge, several methods have been proposed that modify the learned latent space to increase compatibility with the clustering task \cite{madiraju2018deep, xie2016unsupervised}. Generally, this is achieved through the introduction of a complementary loss function, used to increase separability in the latent space. This loss function is referred to as the clustering loss.

\textit{Deep Embedded Clustering (DEC)} is one of the first conventional clustering algorithms presented in this space \cite{xie2016unsupervised}. The proposed time series variant is known as \textit{Deep Temporal Clustering (DTC)} \cite{madiraju2018deep}. Learning in the encoder layers of the DTC model is driven by the interleaved optimisation of two loss functions, namely, the classical reconstruction loss, defined in equation (\ref{eq:1}), as well as the introduced clustering loss, defined in equation (\ref{eq:4}). The DTC model contains a learnable set of cluster centroids. These centroids are used to perform clustering during forward propagation and are updated based on the clustering loss. 

The training procedure used in these models consists of two distinct phases. Firstly, the autoencoder is pretrained to reconstruct the input data. After pretraining is complete, the training data is embedded into the latent space, allowing for the initialisation of the model's cluster centroids. The latent representations are clustered through the use of hierarchical clustering with complete linkage \cite{hubert1974approximate}, and the centroids are initialised as the average of all elements in each cluster. The pretraining phase is used to ensure a meaningful initial set of cluster centroids. In the second phase of training, the clustering loss is introduced. The interleaved optimisation of two loss functions takes place in this phase. The first loss function is the classical reconstruction loss between the input and reconstructed representations, defined in equation (\ref{eq:1}). The second loss is the clustering loss. In each forward pass, the distance from the latent representation $z_i$ to each cluster centroid $\omega_j$ is calculated as $d_{ij}(z_i, \omega_j)$, where $d: \mathcal{Z} \times \mathcal{Z} \rightarrow \mathbb R$ is the distance metric used in the space and $\mathcal{Z}$ is the set of all latent representations. Once calculated, the distances are then normalized to probability assignments $q_{ij}$ using the Student's t-distribution kernel:
\[
    q_{ij} = \frac{(1 + \frac{d(z_i, \omega_j)}{\alpha})^{\frac{-\alpha + 1}{2}}}
    {\sum\limits_{j=1}^{k} \bigg(1 + \frac{d(z_i, \omega_j)}{\alpha} \bigg)^{\frac{-\alpha + 1}{2}}}
\]
where $q_{ij}$ is the probability input $i$ belongs to cluster $j$, $\alpha$ is the number of degrees of freedom, set to one by convention and $k$ is the number of clusters. The clustering loss is formulated as the Kullback–Leibler divergence between the assignment probabilities $q_{ij}$ and the target distribution $p_{ij}$. The target distribution $P$, used in the loss function, is chosen to strengthen high-confidence predictions and normalise losses to prevent distortion of the latent representations:
\[
    p_{ij} = \frac{\frac{q_{ij}^2}{\sum\limits_{i=1}^n q_{ij}}}{\sum\limits_{j=1}^k \frac{q_{ij}^2}{\sum\limits_{i=1}^n q_{ij}}}
\]
This gives the overall loss used in the second phase of training:
\begin{equation} \label{eq:4}
    \mathcal{L_{D}} = \frac{1}{N} \sum\limits_{i=1}^{N} \vert\vert x_i - D_{\theta}(E_{\phi}(x_i)) \vert\vert_2^2 + D_{KL}(Q||P)
\end{equation}
where $Q$ is the cluster assignment distribution, $P$ is the target distribution and $ D_{KL}$ is the Kullback–Leibler divergence.

As the latent representations produced by the DTC encoder are time series as opposed to vectors, the \textit{Complexity Invariant Distance (CID)} is commonly used as a metric function for the latent space. We use $\mathcal{L_{DE}}$ to denote the loss function in which Euclidean distance is used as the metric function and similarly, $\mathcal{L_{DC}}$ for the case where CID is used as the metric function.

\subsection{Dimensionality Reduction}
Dimensionality reduction is the transformation of data from a high-dimensional space to a lower-dimensional space. The objective of dimensionality reduction techniques is to maximally preserve the meaningful properties of the original data after transformation. Dimensionality reduction techniques are commonly divided into linear and non-linear approaches.

\subsubsection{Principal Component Analysis (PCA)}
Principal component analysis is the most widely used linear dimensionality reduction technique. PCA maps the data to a lower-dimensional space in such a way that the variance of the data in the low-dimensional representation is maximized \cite{hotelling1933analysis}. 

\subsubsection{Uniform Manifold Approximation and Projection (UMAP)}
The UMAP algorithm is a general-purpose non-linear dimensionality reduction technique that utilises a theoretical framework based on Riemannian geometry and algebraic
topology \cite{mcinnes2018umap}. 

\section{Methodology}
In this paper, we decompose deep representation learning-based clustering methods into their constituent components before performing an in-depth performance analysis. The objective of this analysis is to find associations between architectural components and good clustering performance when applied in the financial domain. Finally, guided by the results of our performance analysis, we present a novel deep representation learning-based clustering algorithm \textit{Financial Transaction History Clustering (FTHC)}. This algorithm is compared to the current state-of-the-art approaches in the field.

\subsection{Component Selection}
Deep representation learning-based methods are decomposed into four distinct component classes. These component classes consist of (i) the autoencoder architecture, (ii) the dimensionality reduction technique, (iii) the pretext loss function and (iv) the clustering loss function. The full set of components considered is listed in Table \ref{tab:comp1} with more detailed configuration information provided in Appendix A.

\begin{table}[ht]
  \centering
  \begin{tabular}{|m{1.5cm}|c|c|c|c|}
    \hline
    \textbf{\tiny{Component}} & \textbf{\tiny{Option 1}} & \textbf{\tiny{Option 2}} & \textbf{\tiny{Option 3}} & \textbf{\tiny{Option 4}} \\
    \hline
    \tiny{Architecture} & \tiny{FCNN} & \tiny{CNN} & \tiny{LSTM} & \tiny{DTC} \\
    \hline
    \tiny{Dimensionality Reduction} & \tiny{PCA} & \tiny{UMAP} & \tiny{NONE} & \cellcolor{black!50} \\
    \hline
    \tiny{Pretext Loss} & \tiny{$\mathcal{L}_{\mathcal{R}}$} & \tiny{$\mathcal{L}_{\mathcal{LR}}$} & \tiny{$\mathcal{L}_{\mathcal{V}}$} & \tiny{NONE} \\
    \hline
    \tiny{Clustering Loss} & \tiny{$\mathcal{L}_{\mathcal{DE}}$} & \tiny{$\mathcal{L}_{\mathcal{DC}}$} & \tiny{NONE} & \cellcolor{black!50}\\
    \hline
  \end{tabular}
  \caption{Component Configurations}
  \label{tab:comp1}
\end{table}

\subsection{Component Combinations}
A component combination is defined as a set of components including exactly one component from each of the four classes. The performance of each compatible combination is evaluated. A combination is defined as compatible if the components within the combination can function without the violation of any assumptions. For example, the DTC autoencoder architecture is incompatible with the variational autoencoder loss, equation (\ref{eq:3}), as the latent representation produced by the encoder is a time series. Several component incompatibilities exist between architectures and loss functions. Loss function support is described in Table \ref{tab:comp2}. 

\begin{table}[ht]
    \centering
    \begin{tabular}{ |c|c|c|c|c|c|c| }
        \hline
        \multirow{8}{*}{\rotatebox[origin=c]{90}{\tiny{Autoencoder}}} & & \multicolumn{3}{|c|}{\tiny{Pretext Loss}} & \multicolumn{2}{|c|}{\tiny{Clustering Loss}}\\
        \hline
        & & \tiny{$\mathcal{L}_{\mathcal{R}}$} & \tiny{$\mathcal{L}_{\mathcal{LR}}$} & \tiny{$\mathcal{L}_{\mathcal{V}}$} & \tiny{$\mathcal{L}_{\mathcal{DE}}$} & \tiny{$\mathcal{L}_{\mathcal{DC}}$} \\ 
        \hline
        & \tiny{\textit{FCNN}} & \check & \check & \check & \check & \check \\ 
        & \tiny{\textit{CNN}} & \check & \check & \check & \check & \check \\ 
        & \tiny{\textit{LSTM}} & \check & \cross & \check & \check & \check \\ 
        & \tiny{\textit{DTC}} & \check & \cross & \cross & \check & \check\\ 
        \hline
    \end{tabular}
    \caption{Architecture compatibility with loss functions. The LSTM and DTC architectures are incompatible with the extended reconstruction loss, equation (\ref{eq:2}), as the decoders are structured differently from the encoders. This prevents the calculation of the layerwise reconstruction error required by the loss. The DTC architecture is incompatible with the variational autoencoder loss, equation (\ref{eq:3}), as the latent representation produced by the encoder is a time series.}
    \label{tab:comp2}
\end{table}

\subsection{Dataset}
The Berka\footnote{https://data.world/lpetrocelli/czech-financial-dataset-real-anonymized-transactions} dataset consists of a collection of financial records from a bank in Czechoslovakia \cite{berka2000guide}. Specifically, this dataset is an aggregation of the financial information of approximately 5,300 bank customers. The dataset includes over 1,000,000 transactions, close to 700 loans and nearly 900 credit card applications.

\subsection{Evaluation Procedure}
The training procedure\footnote{Training details included in Appendix B.} used for deep representation learning-based models consists of two distinct phases. In the first phase, pretraining, the autoencoder of the model is trained to minimise the pretext loss function. This is followed by the cluster optimisation phase to improve cluster assignments through the minimisation of the clustering loss function. All compatible component combinations were tested.

The dataset was randomly partitioned to form non-overlapping training and testing datasets. Each dataset was composed of 50\% of the available data. Component combinations were trained using the training dataset and evaluated based on the clustering produced on the testing dataset. As described by \citet{ma2019learning, xiao2021rtfn}, although clustering is an unsupervised learning technique, combinations must be evaluated on unseen data to ensure that the learned latent spaces can generalise. Clustering performance is measured in terms of the Silhouette Coefficient and the Davies-Bouldin index. The final performance scores associated with each combination were calculated as the average score achieved across five independent trials.

\section{Results}
In this section, we discuss the results\footnote{Source code and results available at \url{https://github.com/TristanBester/berka_clustering}} obtained from the various analyses carried out in this study. We begin with a discussion relating to the stability of models which incorporate a clustering loss. This is followed by the introduction of a training heuristic for this class of models. Next, we present the performance results for each component class. Finally, we show that our formulation outperforms the current state-of-the-art methods in the domain. 

\subsection{Model Stability}
\begin{figure}[ht]
    \centering
    \includegraphics[width=0.4\textwidth]{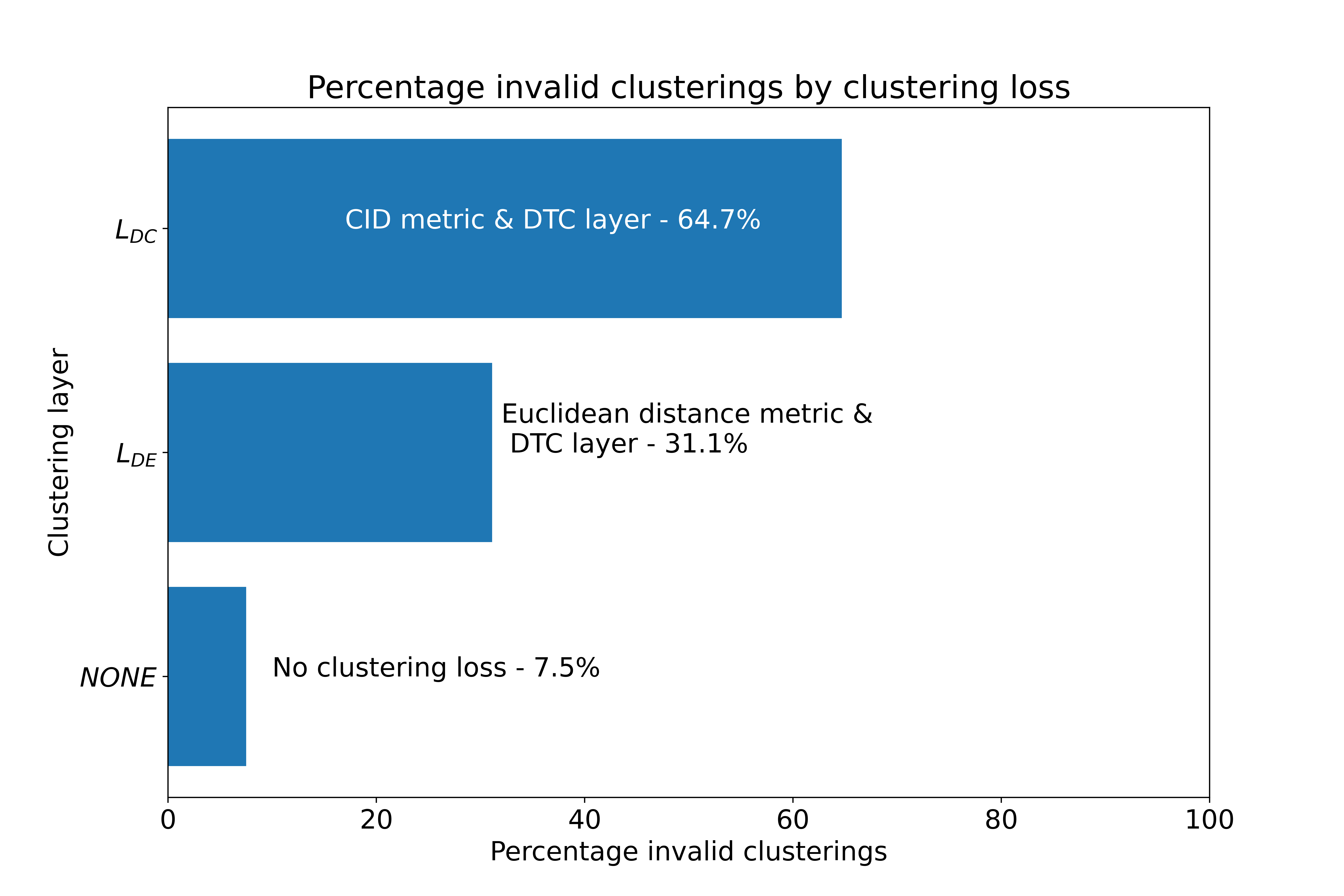}
    \caption{Percentage of invalid clusterings produced across all combinations. Results are shown for each clustering layer variant.}
    \label{fig:inv}
\end{figure}

Throughout the development of the experimental framework, it was observed that invalid clusterings were produced significantly more frequently by certain component combinations than others. A clustering was deemed invalid if one of the following two cases occurred: a degenerate cluster was produced containing all data points or the model was unable to produce cluster assignments as a result of divergence during the training process. In the first case both the SC and DBI are undefined. In the second case, the clusters cannot be computed as a result of numerical errors. The model cannot be evaluated if either of the cases occurs. Upon investigation, it was determined that the only common element between the affected combinations was the inclusion of a clustering loss. This divergent behaviour has previously been attributed to vanishing or exploding gradients \cite{lafabregue2022end}. For combinations that do not incorporate a clustering loss component, we agree with these findings. However, the majority of invalid clusterings are produced by combinations that incorporate a clustering loss as shown in Figure \ref{fig:inv}.

An investigation was carried out to determine the factors leading to the observed behaviour. The clustering loss component used in this study is that of the DTC model proposed by \citet{madiraju2018deep}. To ensure the observed behaviour was not a consequence of the combination of disparate model components, the investigation was carried out using the exact DTC architecture described in the original paper. This restriction would verify that the behaviour was a property of the clustering layer rather than a consequence of the model combination strategy.

An investigation was conducted through the use of a synthetic dataset to test the relationship between the learning rates used in the pretraining and cluster optimisation phases. The learning rate used in the pretraining phase is that which is used to pretrain the autoencoder before cluster optimisation. Similarly, the learning rate used in the cluster optimisation phase is that which is used to minimise the clustering loss function. In the first stage of the investigation, the DTC autoencoder was pretrained on the experimental dataset. After this stage had been completed, the cluster centroids required to initialise the clustering layer are calculated as previously described. As the DTC autoencoder is being used for the experiment, both the latent representations and cluster centroids are time series with a reduced number of time steps. At this point, the autoencoder parameters and cluster centroids were fixed. This allowed for the initial state of the clustering layer to be consistent across all subsequent experiments. Following this step, the learning rates used to train the clustering layers were varied and the results were recorded for analysis. 

Polynomials of varying degrees were used to generate synthetic time series for the experimental dataset.  Cluster labels were assigned to each of the time series in the dataset based on the degree of the polynomial from which it was generated. The pretraining phase was carried out on the dataset, followed by the centroid initialisation process. The latent space produced by the encoder at this point is illustrated in Appendix C. A learning rate of $\eta_{pre}$ was used in the pretraining phase and $\eta_{cls}$ in the cluster optimisation phase. The effect of varying the learning rate used in the cluster optimisation phase is illustrated in Figure 2. 
\begin{figure}[ht]
    \centering
    \includegraphics[width=0.4\textwidth]{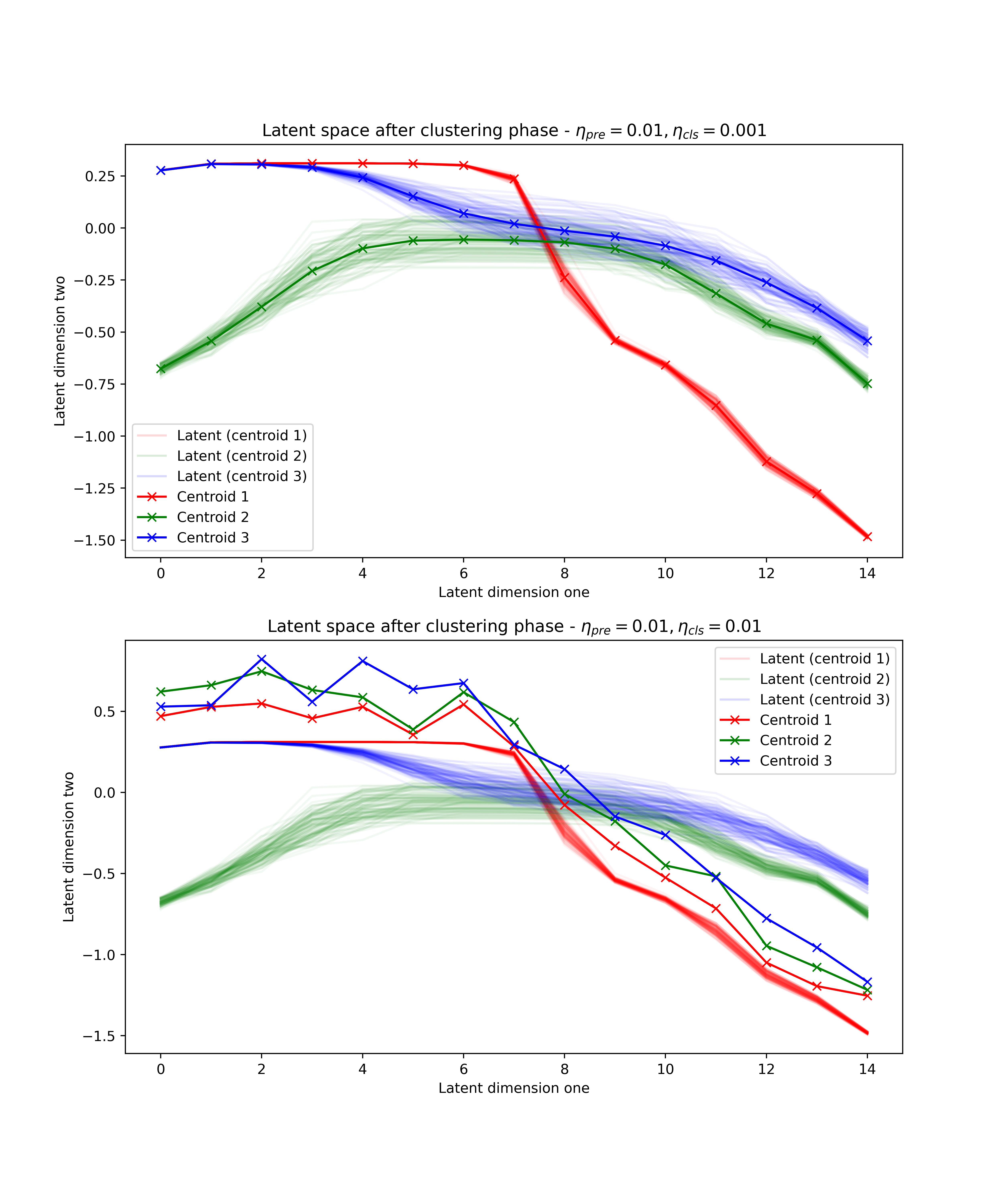}
    \caption{The effect of varied learning rates in the cluster optimisation phase. In the lower plot, it can be seen that the cluster centroids remain in their initial positions while the latent representations all converge to a single representation. Consequently, all data points are assigned to the same cluster. The stability of the upper model is clear from the converged latent space representation.\protect\footnotemark}
    \label{fig:conv2}
\end{figure}
\footnotetext{The loss curve of the upper model flattened out towards the end of training indicating that the model had converged to a stable solution rather than requiring more iterations to diverge as a result of the smaller learning rate.}

After analysis of the recorded results, a subset of which have been presented, the cause for the behaviour was identified. If the learning rate used in the cluster optimisation phase was greater than or equal to that of the pretraining phase, the model was significantly more likely to produce an invalid clustering. Moreover, it was observed that model stability could be significantly improved by setting the learning rate used in the cluster optimisation phase an order of magnitude smaller than that used in the pretraining phase.

\subsection{Performance Results}
An analysis was conducted to assess whether or not clusters separate consumers based on human-understandable properties of transaction histories. It was found that in general clusters associate consumers with similar transaction behaviour (i.e. salaried employees are separated from individuals with irregular income) however counterexamples exist within the clusters. As a result, we feel that research into human-interpretable performance metrics for transaction history clustering is an interesting avenue for future work. 

From the results, it is clear that no single combination outperforms all others in each test. That is, for each performance metric and number of clusters, there is no single best algorithm. However, certain combinations are more commonly associated with higher performance than others. The performance associated with each component is calculated as the average performance achieved by all combinations in which the component is used. The results obtained for each component class are described below.

\subsubsection{Autoencoder Architecture}
\begin{figure}[ht]
    \centering
    \includegraphics[width=0.4\textwidth]{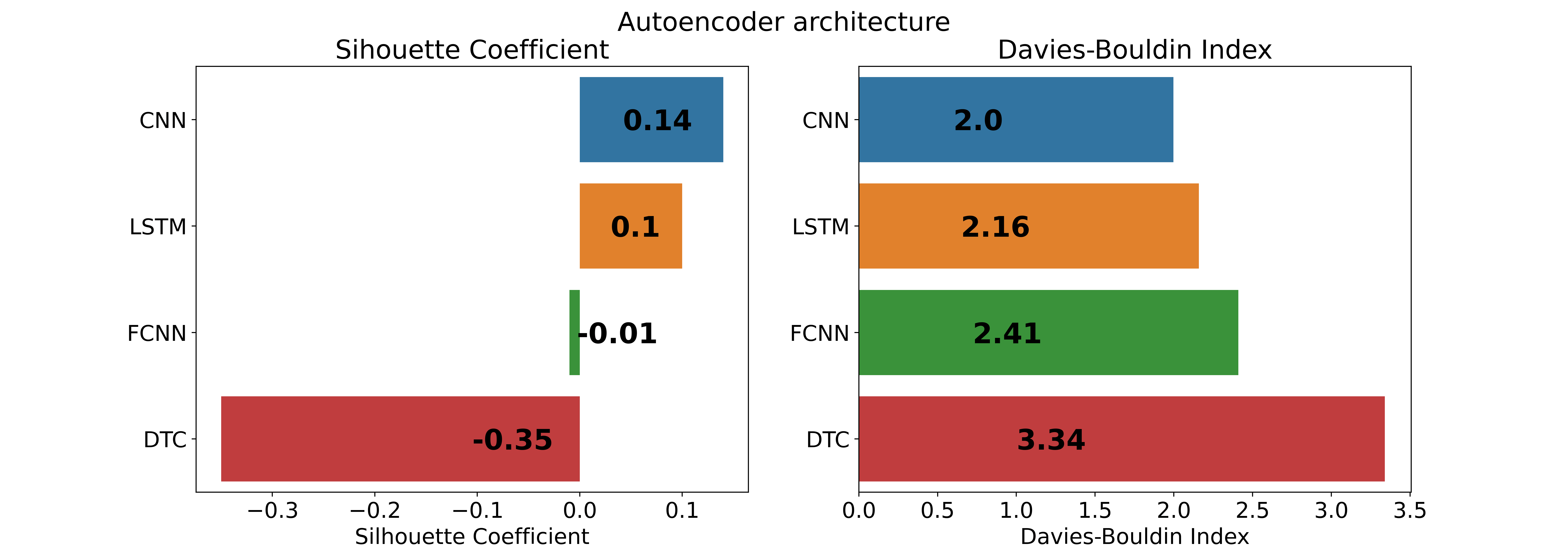}
    \caption{Average clustering performance associated with each autoencoder architecture.}
    \label{fig:ae}
\end{figure}

The results obtained for the autoencoder architecture component are consistent across both performance metrics. This is illustrated in Figure \ref{fig:ae}. For this component class, the CNN-based component is associated with the highest performance. This is consistent with the results obtained by \citet{lafabregue2022end} on the univariate UCR Archive\footnote{https://www.cs.ucr.edu/$\sim$eamonn/time\_series\_data}.

\subsubsection{Pretext Loss Function}
\begin{figure}[ht]
    \centering
    \includegraphics[width=0.4\textwidth]{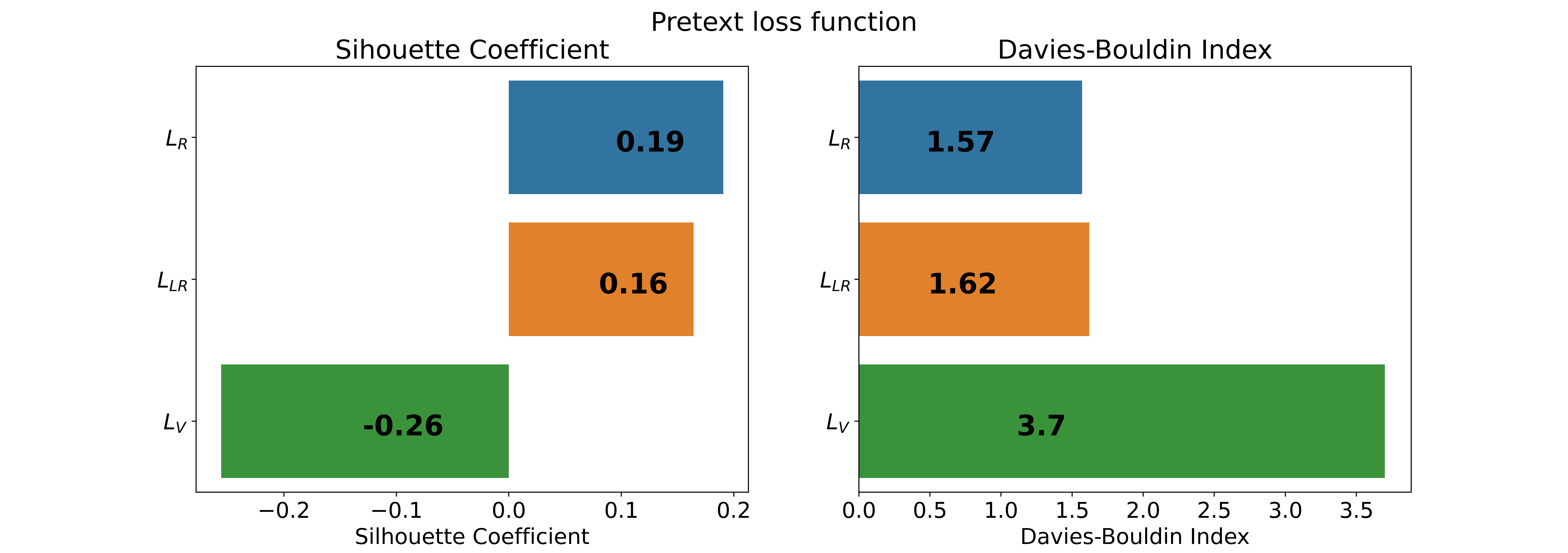}
    \caption{Average clustering performance associated with each pretext loss function.}
    \label{fig:pre}
\end{figure}

The classical reconstruction loss, equation (\ref{eq:1}), is associated with the highest average performance across all combinations. The results are illustrated in Figure \ref{fig:pre}. Our results are in agreement with \citet{lafabregue2022end} in which it is stated that the representations learned for the generation task, equation (\ref{eq:3}), are incompatible with clustering.

\subsubsection{Clustering Loss Function}
\begin{figure}[ht]
    \centering
    \includegraphics[width=0.4\textwidth]{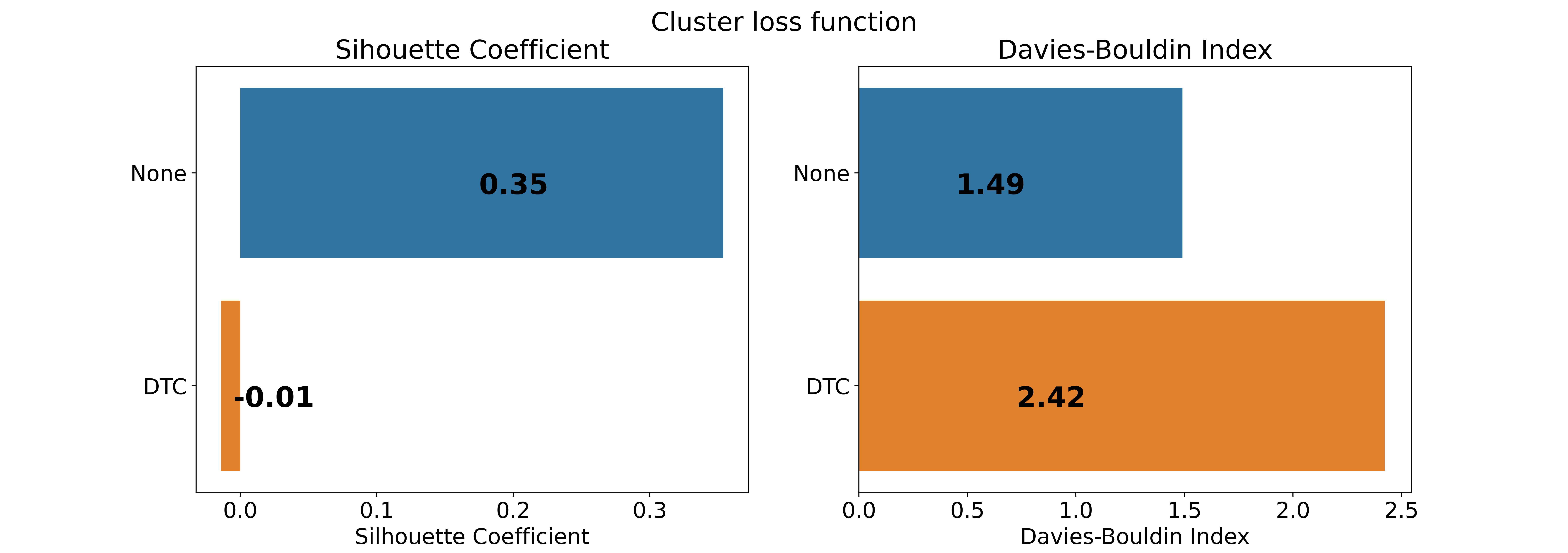}
    \caption{Average clustering performance associated with each clustering loss function.}
    \label{fig:clust}
\end{figure}

Higher performance is associated with combinations in which a clustering loss is not used. As illustrated in Figure \ref{fig:clust}, a significant increase in average performance is obtained with the removal of the clustering loss. Contrary to the results presented by \citet{lafabregue2022end}, $28$ out of the $30$ highest performance combinations made use of a clustering loss. It was previously concluded that the use of existing clustering losses is not relevant for time series. This is likely a result of the fact that the learning rate used in the clustering phase was not tailored to increase model stability. Consequently, the models performed poorly or produced invalid clusterings. We rely on these results in the construction of the FTHC formulation. That is when a stable solution is obtained with the use of a clustering loss, the combination will perform better than if the component had been omitted.

\subsubsection{Clustering Distance Metrics}
\begin{figure}[ht]
    \centering
    \includegraphics[width=0.4\textwidth]{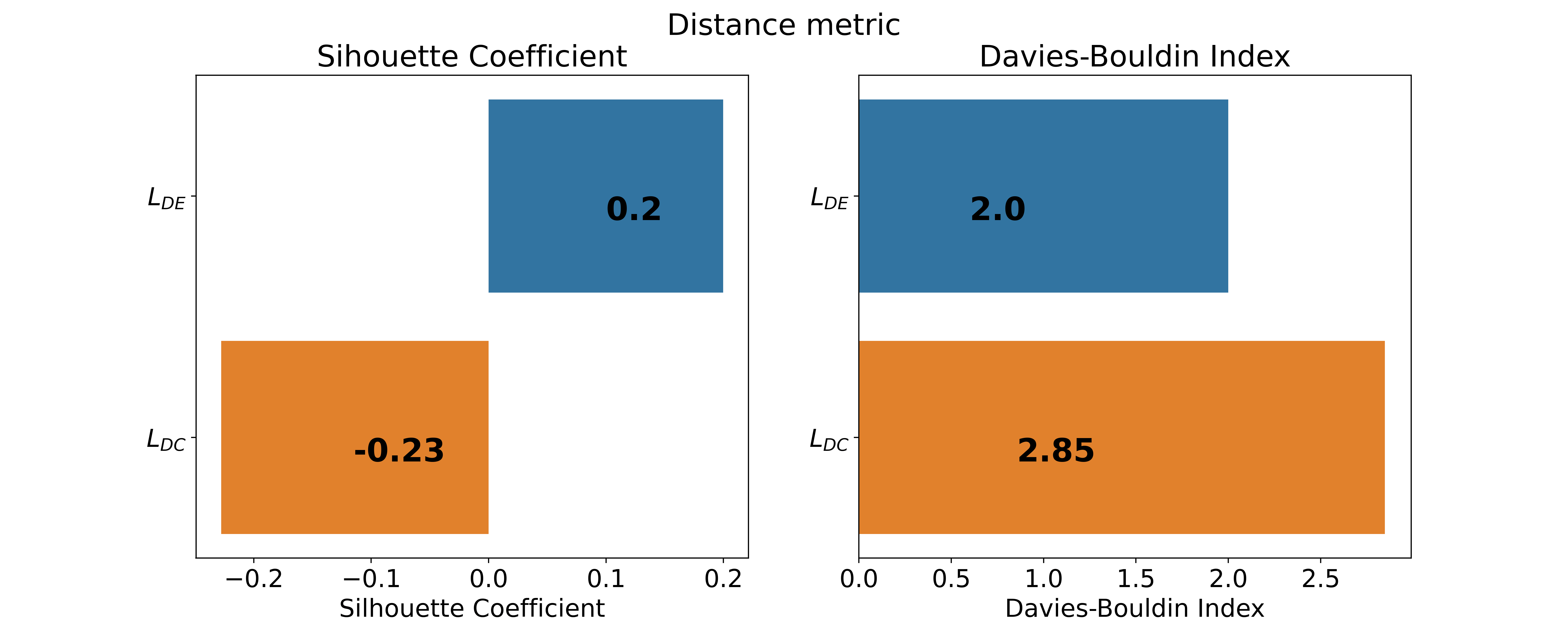}
    \caption{Average performance associated with clustering loss function metrics.}
    \label{fig:dist}
\end{figure}

Of the two functions used as metrics in the latent space, the Euclidean distance was associated with the highest performance. On average when Euclidean distance was used as the clustering layer metric function, the performance was significantly higher than that associated with complexity invariant distance. These results are shown in Figure \ref{fig:dist}.

\subsection{Dimensionality reduction}
\begin{figure}[ht]
    \centering
    \includegraphics[width=0.4\textwidth]{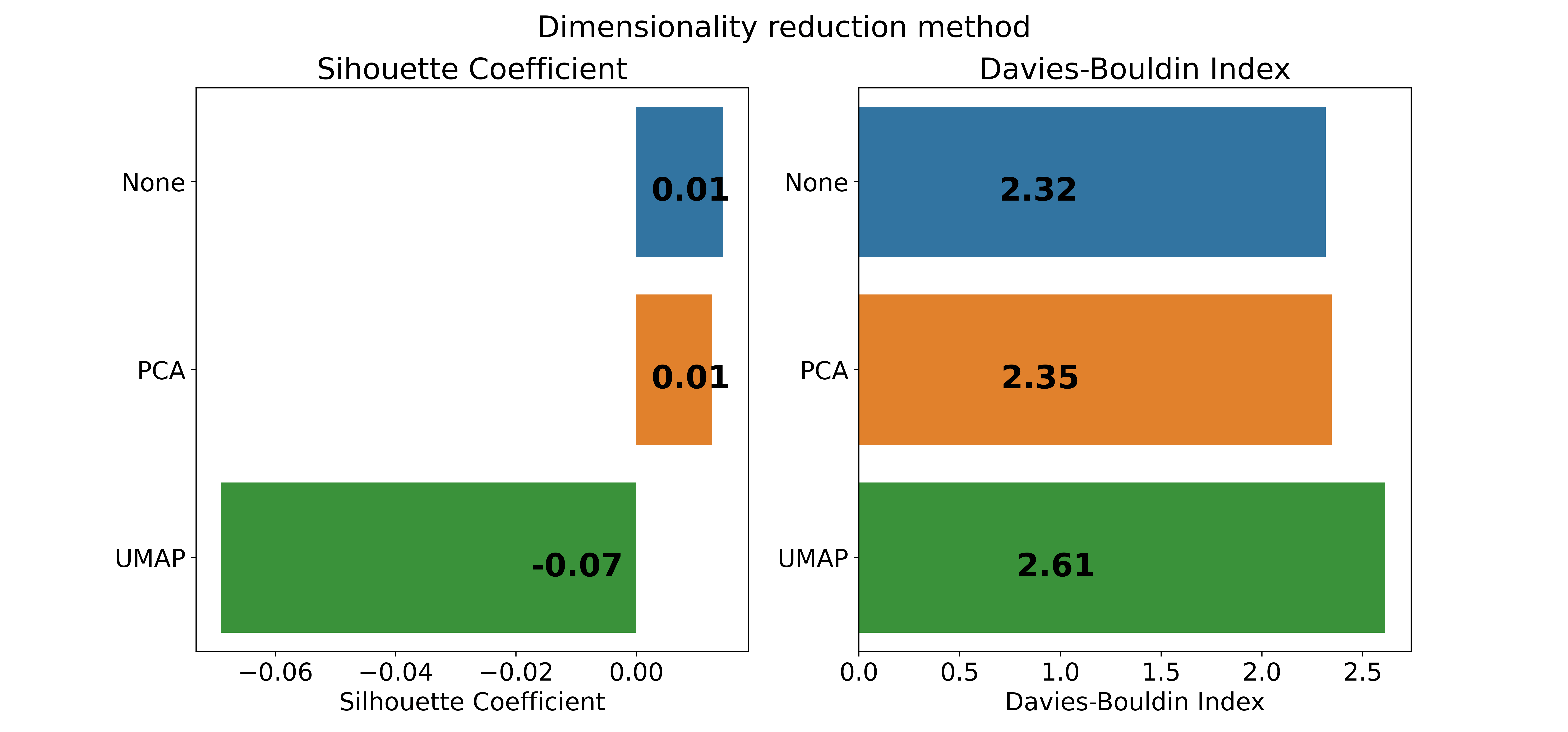}
    \caption{Average clustering performance associated with dimensionality reduction techniques.}
    \label{fig:dim}
\end{figure}

\noindent
Across all component combinations, principal component analysis and the omission of dimensionality reduction exhibit similar average performance characteristics. The incorporation of UMAP as a dimensionality reduction technique shows slightly decreased performance as illustrated in Figure \ref{fig:dim}.

\subsection{Financial Transaction History Clustering (FTHC)}
\begin{figure}[ht]
    \centering
    \includegraphics[width=0.5\textwidth]{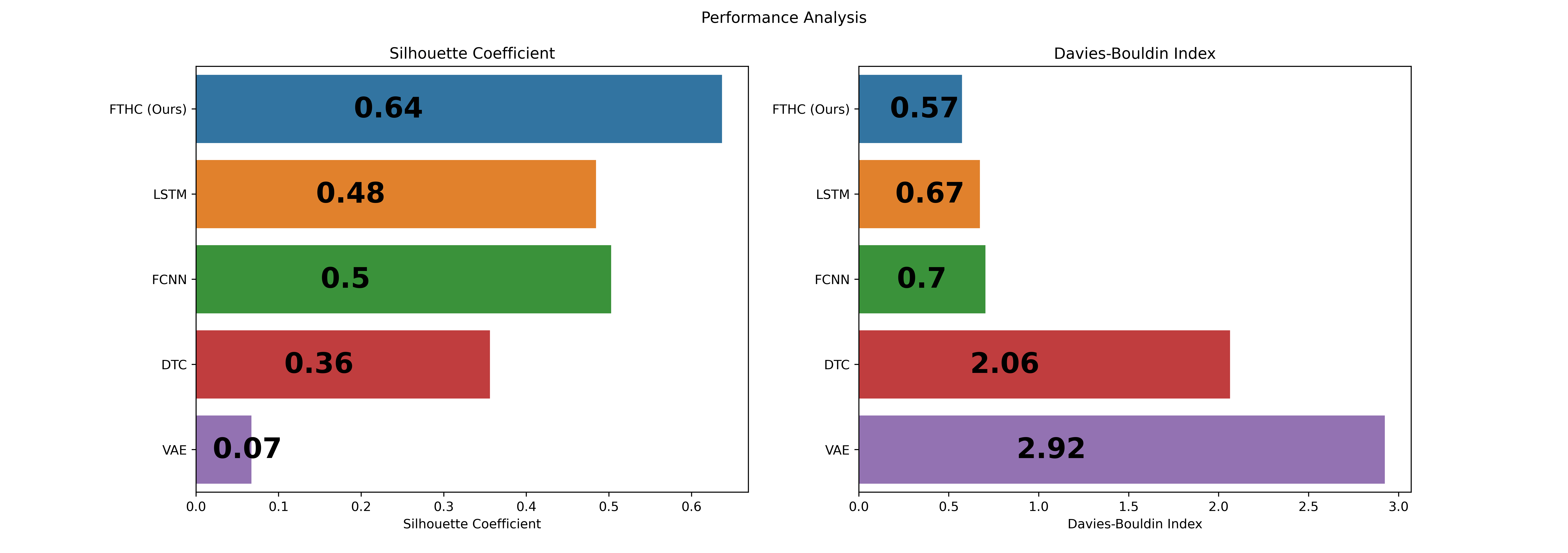}
    \caption{Performance results comparing our approach to the current state-of-the-art methods. The K-means clustering algorithm is used to form clusters in the latent space of the \textit{LSTM, FCNN} and \textit{VAE} autoencoders. The \textit{DTC} approach forms clusters through the use of a clustering layer. Results were averaged across five independent trials.}
    \label{fig:fthc}
\end{figure}

In this section, we define a novel approach designed for financial time series clustering before establishing a strong performance baseline for the domain. Guided by the previously discussed results, we combine the following components to produce a novel approach known as \textit{Financial Transaction History Clustering}. We make use of the CNN-based autoencoder architecture based on ResNet. The autoencoder is trained using the classical reconstruction loss as a pretext loss function. The encoder is trained with the DTC clustering loss function with Euclidean distance as the metric. We compare our approach to the current state-of-the-art techniques in financial time series clustering presented by \citet{lafabregue2022end, barkhordar2021clustering}. From the results presented in Figure \ref{fig:fthc}, it can be seen that FTHC outperforms all of the current state-of-the-art approaches both in terms of the SC as well as the DBI.

\section{Conclusion}
In this paper, we have conducted a study to compare a variety of time series clustering algorithms based on deep representation learning. The algorithms are compared based on their ability to cluster univariate time series of consumer bank transaction data. We have shown that deep learning-based clustering methods can be decomposed into four components, namely, the architecture, dimensionality reduction technique, pretext loss function and clustering loss function. Based on the proposed taxonomy,  we have conducted a cross-comparison to evaluate each component's influence on clustering performance. We combine the top-performing components to create a novel clustering algorithm which is shown to outperform the current state-of-the-art approaches both in terms of the silhouette coefficient as well as the Davies-Bouldin index.

\bibliography{aaai24}

\begin{thebibliography}{27}
\providecommand{\natexlab}[1]{#1}

\bibitem[{Alshehadeh and Al-Khawaja(2022)}]{alshehadeh2022financial}
Alshehadeh, A.~R.; and Al-Khawaja, H.~A. 2022.
\newblock Financial Technology as a Basis for Financial Inclusion and its Impact on Profitability: Evidence from Commercial Banks.
\newblock \emph{Int. J. Advance Soft Compu. Appl}, 14(2).

\bibitem[{Ansari, Riasi et~al.(2016)}]{ansari2016customer}
Ansari, A.; Riasi, A.; et~al. 2016.
\newblock Customer clustering using a combination of fuzzy c-means and genetic algorithms.
\newblock \emph{International Journal of Business and Management}, 11(7): 59--66.

\bibitem[{Barkhordar, Shirali-Shahreza, and Sadeghi(2021)}]{barkhordar2021clustering}
Barkhordar, E.; Shirali-Shahreza, M.~H.; and Sadeghi, H.~R. 2021.
\newblock Clustering of Bank Customers using LSTM-based encoder-decoder and Dynamic Time Warping.
\newblock \emph{arXiv preprint arXiv:2110.11769}.

\bibitem[{Berka et~al.(2000)}]{berka2000guide}
Berka, P.; et~al. 2000.
\newblock Guide to the financial data set.
\newblock \emph{PKDD2000 discovery challenge}.

\bibitem[{Block, Knight~Jr, and Rosenblatt(1962)}]{block1962analysis}
Block, H.~D.; Knight~Jr, B.; and Rosenblatt, F. 1962.
\newblock Analysis of a four-layer series-coupled perceptron. II.
\newblock \emph{Reviews of Modern Physics}, 34(1): 135.

\bibitem[{Dempster, Petitjean, and Webb(2020)}]{dempster2020rocket}
Dempster, A.; Petitjean, F.; and Webb, G.~I. 2020.
\newblock ROCKET: exceptionally fast and accurate time series classification using random convolutional kernels.
\newblock \emph{Data Mining and Knowledge Discovery}, 34(5): 1454--1495.

\bibitem[{Ghasedi~Dizaji et~al.(2017)Ghasedi~Dizaji, Herandi, Deng, Cai, and Huang}]{ghasedi2017deep}
Ghasedi~Dizaji, K.; Herandi, A.; Deng, C.; Cai, W.; and Huang, H. 2017.
\newblock Deep clustering via joint convolutional autoencoder embedding and relative entropy minimization.
\newblock In \emph{Proceedings of the IEEE international conference on computer vision}, 5736--5745.

\bibitem[{Guo et~al.(2017)Guo, Gao, Liu, and Yin}]{guo2017improved}
Guo, X.; Gao, L.; Liu, X.; and Yin, J. 2017.
\newblock Improved deep embedded clustering with local structure preservation.
\newblock In \emph{Ijcai}, volume~17, 1753--1759.

\bibitem[{He et~al.(2016)He, Zhang, Ren, and Sun}]{he2016deep}
He, K.; Zhang, X.; Ren, S.; and Sun, J. 2016.
\newblock Deep residual learning for image recognition.
\newblock In \emph{Proceedings of the IEEE conference on computer vision and pattern recognition}, 770--778.

\bibitem[{Hochreiter and Schmidhuber(1997)}]{hochreiter1997long}
Hochreiter, S.; and Schmidhuber, J. 1997.
\newblock Long short-term memory.
\newblock \emph{Neural computation}, 9(8): 1735--1780.

\bibitem[{Hopfield(1982)}]{hopfield1982neural}
Hopfield, J.~J. 1982.
\newblock Neural networks and physical systems with emergent collective computational abilities.
\newblock \emph{Proceedings of the national academy of sciences}, 79(8): 2554--2558.

\bibitem[{Hotelling(1933)}]{hotelling1933analysis}
Hotelling, H. 1933.
\newblock Analysis of a complex of statistical variables into principal components.
\newblock \emph{Journal of educational psychology}, 24(6): 417.

\bibitem[{Hubert(1974)}]{hubert1974approximate}
Hubert, L. 1974.
\newblock Approximate evaluation techniques for the single-link and complete-link hierarchical clustering procedures.
\newblock \emph{Journal of the American Statistical Association}, 69(347): 698--704.

\bibitem[{Ismail~Fawaz et~al.(2019)Ismail~Fawaz, Forestier, Weber, Idoumghar, and Muller}]{ismail2019deep}
Ismail~Fawaz, H.; Forestier, G.; Weber, J.; Idoumghar, L.; and Muller, P.-A. 2019.
\newblock Deep learning for time series classification: a review.
\newblock \emph{Data mining and knowledge discovery}, 33(4): 917--963.

\bibitem[{Kingma and Welling(2013)}]{kingma2013auto}
Kingma, D.~P.; and Welling, M. 2013.
\newblock Auto-encoding variational bayes.
\newblock \emph{arXiv preprint arXiv:1312.6114}.

\bibitem[{Lafabregue et~al.(2022)Lafabregue, Weber, Gan{\c{c}}arski, and Forestier}]{lafabregue2022end}
Lafabregue, B.; Weber, J.; Gan{\c{c}}arski, P.; and Forestier, G. 2022.
\newblock End-to-end deep representation learning for time series clustering: a comparative study.
\newblock \emph{Data Mining and Knowledge Discovery}, 36(1): 29--81.

\bibitem[{Ma et~al.(2019)Ma, Zheng, Li, and Cottrell}]{ma2019learning}
Ma, Q.; Zheng, J.; Li, S.; and Cottrell, G.~W. 2019.
\newblock Learning representations for time series clustering.
\newblock \emph{Advances in neural information processing systems}, 32.

\bibitem[{MacQueen et~al.(1967)}]{macqueen1967some}
MacQueen, J.; et~al. 1967.
\newblock Some methods for classification and analysis of multivariate observations.
\newblock In \emph{Proceedings of the fifth Berkeley symposium on mathematical statistics and probability}, volume~1, 281--297. Oakland, CA, USA.

\bibitem[{Madiraju(2018)}]{madiraju2018deep}
Madiraju, N.~S. 2018.
\newblock \emph{Deep temporal clustering: Fully unsupervised learning of time-domain features}.
\newblock Ph.D. thesis, Arizona State University.

\bibitem[{McInnes, Healy, and Melville(2018)}]{mcinnes2018umap}
McInnes, L.; Healy, J.; and Melville, J. 2018.
\newblock Umap: Uniform manifold approximation and projection for dimension reduction.
\newblock \emph{arXiv preprint arXiv:1802.03426}.

\bibitem[{Meng et~al.(2017)Meng, Catchpoole, Skillicom, and Kennedy}]{meng2017relational}
Meng, Q.; Catchpoole, D.; Skillicom, D.; and Kennedy, P.~J. 2017.
\newblock Relational autoencoder for feature extraction.
\newblock In \emph{2017 International joint conference on neural networks (IJCNN)}, 364--371. IEEE.

\bibitem[{Moin and Ahmed(2012)}]{moin2012use}
Moin, K.~I.; and Ahmed, D. Q.~B. 2012.
\newblock Use of data mining in banking.
\newblock \emph{International Journal of Engineering Research and Applications}, 2(2): 738--742.

\bibitem[{M{\"u}ller(2007)}]{muller2007dynamic}
M{\"u}ller, M. 2007.
\newblock Dynamic time warping.
\newblock \emph{Information retrieval for music and motion}, 69--84.

\bibitem[{Taifi(2023)}]{taifi2023intertwined}
Taifi, N. 2023.
\newblock The Intertwined Role of Big Data with Business Intelligence and Real Time Analysis: A Finance Sector Perspective.
\newblock In \emph{2023 International Conference on Digital Age \& Technological Advances for Sustainable Development (ICDATA)}, 17--23. IEEE.

\bibitem[{Wang, Yan, and Oates(2017)}]{wang2017time}
Wang, Z.; Yan, W.; and Oates, T. 2017.
\newblock Time series classification from scratch with deep neural networks: A strong baseline.
\newblock In \emph{2017 International joint conference on neural networks (IJCNN)}, 1578--1585. IEEE.

\bibitem[{Xiao et~al.(2021)Xiao, Xu, Xing, Luo, Dai, and Zhan}]{xiao2021rtfn}
Xiao, Z.; Xu, X.; Xing, H.; Luo, S.; Dai, P.; and Zhan, D. 2021.
\newblock RTFN: A robust temporal feature network for time series classification.
\newblock \emph{Information sciences}, 571: 65--86.

\bibitem[{Xie, Girshick, and Farhadi(2016)}]{xie2016unsupervised}
Xie, J.; Girshick, R.; and Farhadi, A. 2016.
\newblock Unsupervised deep embedding for clustering analysis.
\newblock In \emph{International conference on machine learning}, 478--487. PMLR.

\end{thebibliography}

\clearpage
\appendix
\section{Appendix}
\subsection{Appendix A: Component Configurations}
\subsubsection{(i) Autoencoder Architecture}
The configuration of each architecture depends on a large number of hyperparameters. As a result, we have decided to use the configurations used in other articles for our experiments as described below.

\begin{itemize}
    \item \textit{FCNN} -- The fully connected neural network architecture proposed in \cite{xie2016unsupervised, guo2017improved} will be used to represent this class of autoencoder architectures. The encoder is composed of three FCNN layers. The number of neurons in each layer is 500, 500 and 2,000 respectively. An embedding layer is used to map the data into the latent space. The decoder is constructed as a mirror of the encoder with the exception of the embedding layer.
    \item \textit{CNN} -- The convolutional neural network architecture is based on the ResNet architecture described in \cite{he2016deep}. The encoder is composed of three residual blocks followed by a global average pooling layer. A fully connected layer is used as the embedding layer. Similar to the FCNN, the decoder is constructed as a mirror of the encoder with the exception of the embedding layer.
    \item \textit{LSTM} -- The first recurrent neural network architecture is based on the LSTM. The encoder is comprised of a stacked, two-layer bidirectional LSTM. The network outputs associated with the forward and backward passes at each time step are concatenated to form the final output. The final hidden state of the network is used as the latent representation. The decoder is constructed as a mirror of the encoder and latent representation is upsampled to ensure that the input and output sequence lengths are identical.
    \item \textit{DTC} -- The second recurrent neural network architecture has been proposed in \citet{madiraju2018deep}. Notably, the latent representation produced by this network is a time series with a reduced number of time steps. The encoder is composed of a convolutional layer followed by a max-pooling layer with a kernel size of 10. The output of this layer is then passed to a stacked, two-layer bidirectional LSTM with a hidden size of 50. Once again, forward and backward pass outputs are concatenated at each time step. The latent representation of the input time series is equal to the hidden state sequence produced by the encoder. The decoder is comprised of an upsampling layer followed by a deconvolutional layer to reconstruct the input time series.
\end{itemize}

\subsubsection{(ii) Dimensionality Reduction}
\begin{itemize}
    \item \textit{PCA} -- Principal component analysis is used to reduce the dimensionality of the latent vectors before the clustering operation.
    \item \textit{UMAP} -- The UMAP algorithm is used to reduce the dimensionality of the latent vectors before clustering.
    \item \textit{NONE} -- In this case, no dimensionality reduction is performed. As a result, the latent representations are clustered directly.
\end{itemize}

\subsubsection{(iii) Pretext Loss Function}
\begin{itemize}
    \item $\mathcal{L}_{\mathcal{R}}$ -- Equation (\ref{eq:1})
    \item $\mathcal{L}_{\mathcal{LR}}$ -- Equation (\ref{eq:2})
    \item $\mathcal{L}_{\mathcal{V}}$ -- Equation (\ref{eq:3})
\end{itemize}

\subsubsection{(iv) Clustering Loss Function}
\begin{itemize}
    \item $\mathcal{L}_{\mathcal{DE}}$ -- This is the DTC clustering loss function in which the Euclidean distance is used as the metric function. That is, the Euclidean distance is used to measure distances in the latent space of the autoencoder.
    \item $\mathcal{L}_{\mathcal{DC}}$ --  This is the DTC clustering loss function in which Complexity Invariant Distance is used as the metric function. That is, the CID is used to measure distances in the latent space of the autoencoder.
    \item \textit{NONE} -- In this case, no clustering loss function is used. As a result, the model is an autoencoder in which clustering is performed in the latent space.
\end{itemize}

\subsection{Appendix B: Model Training Procedure}
All compatible component combinations were tested. Each combination was trained for 1,000 batch iterations in the pretraining phase and 1,000 batch iterations in the cluster optimisation phase. A constant learning rate of $\eta_{pre}=10^{-2}$ was used in the pretraining phase and $\eta_{cls}=10^{-3}$ in the cluster optimisation phase. The selection of these learning rates is motivated in the next section. For combinations in which a clustering loss is not included, training is terminated at the end of the pretraining phase.

 For combinations that do not make use of a clustering layer, clusters are calculated using the K-Means algorithm \cite{macqueen1967some} to cluster the latent representations produced by the model. This is not necessary with the inclusion of a clustering layer, as the layer produces cluster assignments directly.

\subsection{Appendix C: Latent Space After Pretraining}
\begin{figure}[ht]
    \centering
    \includegraphics[width=0.4\textwidth]{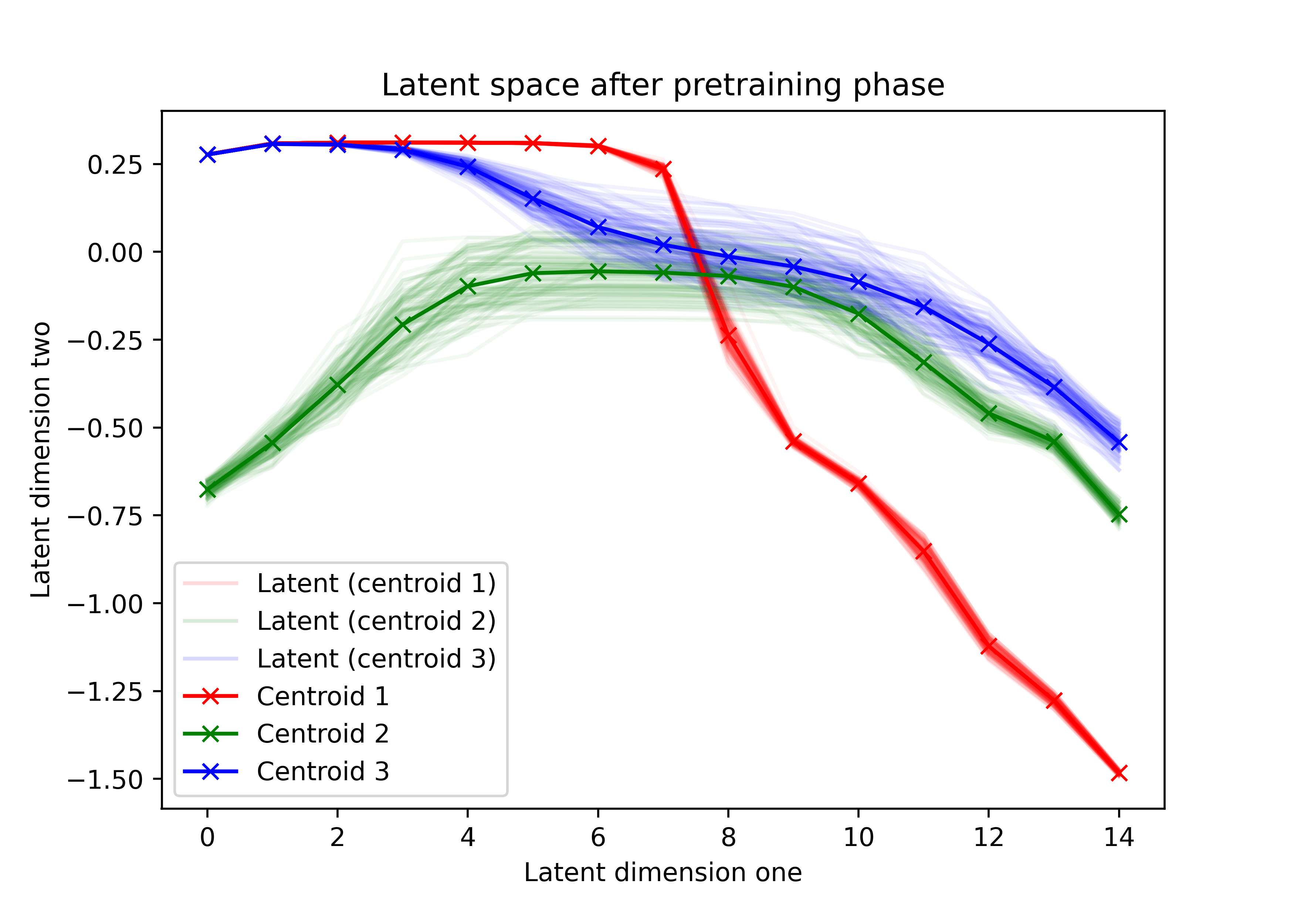}
    \caption{Visualisation of the autoencoder latent space after the pertaining phase.}
    \label{fig:latent}
\end{figure}

\end{document}